\documentclass[11pt,a4paper]{article}
\usepackage[hyperref]{naaclhlt2019}
\usepackage{times}
\usepackage{latexsym}
\usepackage{mathtools}
\usepackage{url}
\usepackage{multirow}
\usepackage{array}
\usepackage{arydshln}
\usepackage{tabularx}
\usepackage{makecell}
\usepackage{graphicx}  
\aclfinalcopy 
\def\aclpaperid{2379} 

\newcounter{Lcount}
\newcommand{\squishenum}{
\begin{list}{\arabic{Lcount}. }
{ \usecounter{Lcount}
\setlength{\itemsep}{0pt}
\setlength{\parsep}{0pt}
\setlength{\topsep}{0pt}
\setlength{\partopsep}{0pt}
\setlength{\leftmargin}{2em}
\setlength{\labelwidth}{1.5em}
\setlength{\labelsep}{0.5em} } }

\newcommand{\squishletter}{
\begin{list}{\alph{Lcount}. }
{ \usecounter{Lcount}
	\setlength{\itemsep}{0pt}
	\setlength{\parsep}{0pt}
	\setlength{\topsep}{0pt}
	\setlength{\partopsep}{0pt}
	\setlength{\leftmargin}{2em}
	\setlength{\labelwidth}{1.5em}
	\setlength{\labelsep}{0.5em} } }

\newcommand{\squishlist}{
\begin{list}{$\bullet$}
	{ \usecounter{Lcount}
		\setlength{\itemsep}{0pt}
		\setlength{\parsep}{0pt}
		\setlength{\topsep}{0pt}
		\setlength{\partopsep}{0pt}
		\setlength{\leftmargin}{2em}
		\setlength{\labelwidth}{1.5em}
		\setlength{\labelsep}{0.5em} } }

\newcommand{\squishend}{
\end{list} }

\title{Joint Detection and Location of English Puns}

\author{%
Yanyan Zou \and Wei Lu\\
StatNLP Research Group\\
Singapore University of Technology and Design \\
{\tt yanyan\_zou@mymail.sutd.edu.sg, luwei@sutd.edu.sg} \\
}

\date{}

\begin{document}
\maketitle
\begin{abstract}
	A pun is a form of wordplay for an intended humorous or rhetorical effect, where a word suggests two or more meanings by exploiting polysemy (\emph{homographic pun}) or phonological similarity to another word (\emph{heterographic pun}).
	This paper presents an approach that addresses pun detection and pun location jointly from a sequence labeling perspective.
	We employ a new tagging scheme such that the model is capable of performing such a joint task, where useful structural information can be properly captured.
We show that our proposed model is effective in handling both homographic and heterographic puns.
	Empirical results on the benchmark datasets demonstrate that our approach can achieve new state-of-the-art results.
\end{abstract}

\section{Introduction}
There exists a class of language construction known as \emph{pun} in natural language texts and utterances, where a certain word or other lexical items are used to exploit two or more separate meanings.
It has been shown that understanding of puns is an important research question with various real-world applications, such as human-computer interaction \cite{morkes1999effects,hempelmann2008computational}
and machine translation \cite{schroter2005shun}.
Recently, many researchers show their interests in studying puns, like detecting pun sentences \cite{vadehra2017uwav}, locating puns in the text \cite{cai2018sense}, interpreting pun sentences \cite{sevgili2017n} and generating sentences containing puns \cite{ritchie2005computational,hong2009automatically,wan2018neural}.
A pun is a wordplay in which a certain word suggests two or more meanings by exploiting polysemy, homonymy, or phonological similarity to another sign, for an intended humorous or rhetorical effect.
Puns can be generally categorized into two groups, namely heterographic puns (where the pun and its latent target are phonologically similar) and homographic puns (where the two meanings of the pun reflect its two distinct senses) \cite{miller2017semeval}.  
Consider the following two examples:

\begin{tabular}{cp{6cm}}
	\vspace{2mm}
	(1) &	When the church bought gas for their annual barbecue, proceeds went from the sacred to the \emph{propane}. \\
	(2)	& Some diets cause a {\em gut} reaction. \\
\end{tabular}

\noindent
The first punning joke exploits the sound similarity between the word ``\emph{propane}" and the latent target ``profane", which can be categorized into the group of heterographic puns.
Another categorization of English puns is homographic pun, exemplified by the second instance leveraging distinct senses of the word ``\emph{gut}".

Pun detection is the task of detecting whether there is a pun residing in the given text.
The goal of pun location is to find the exact word appearing in the text that implies more than one meanings.
Most previous work addresses such two tasks  separately and develop separate systems \cite{pramanick2017ju,sevgili2017n}.
Typically, a system for pun detection is built to make a binary prediction on whether a sentence contains a pun or not, where all instances (with or without puns) are taken into account during training.
For the task of pun location, a separate system is used to make a single prediction as to which word in the given sentence in the text that trigger more than one semantic interpretations of the text, where the training data involves only sentences that contain a pun.
Therefore, if one is interested in solving both problems at the same time, a pipeline approach that performs pun detection followed by pun location can be used.

Compared to the pipeline methods, joint learning has been shown effective \cite{katiyar2016investigating,peng2018learning} since it is able to reduce error propagation and allows information exchange between tasks which is potentially beneficial to all the tasks. 
In this work, we demonstrate that the detection and location of puns can be jointly addressed by a single model. 
The pun detection and location tasks can be combined as a sequence labeling problem, which allows us to jointly detect and locate a pun in a sentence by assigning each word a tag.
Since each context contains a maximum of one pun \cite{miller2017semeval}, we design a novel tagging scheme to capture this structural constraint.
Statistics on the corpora also show that a pun tends to appear in the second half of a context.
To capture such a structural property, we also incorporate word position knowledge into our structured prediction model.
Experiments on the benchmark datasets show that detection and location tasks can reinforce each other, leading to new state-of-the-art performance on these two tasks.
To the best of our knowledge, this is the first work that performs joint detection and location of English puns by using a sequence labeling approach.\footnote{
Our code is publicly available at \url{https://github.com/zoezou2015/PunLocation}.}


\section{Approach}

\subsection{Problem Definition}

We first design a simple tagging scheme consisting of two tags \{$\mathsf{N, P}$\}:
\squishlist
\item $\mathsf{N}$ tag means the current word is \underline{n}ot a pun.
\item $\mathsf{P}$ tag means the current word is a \underline{p}un.
\squishend
If the tag sequence of a sentence contains a $\mathsf{P}$ tag, then the text contains a pun and the word corresponding to $\mathsf{P}$ is the pun.

The contexts have the characteristic that each context contains a maximum of one pun \cite{miller2017semeval}.
In other words, there exists only one pun if the given sentence is detected as the one containing a pun.
Otherwise, there is no pun residing in the text.
To capture this interesting property, we propose a new tagging scheme consisting of three tags, namely \{$\mathsf{B, P, A}$\}.

\squishlist
\item $\mathsf{B}$ tag indicates that the current word appears \underline{b}efore the pun in the given context.
\item $\mathsf{P}$ tag highlights the current word is a \underline{p}un.
\item $\mathsf{A}$ tag indicates that the current word appears \underline{a}fter the pun. 
\squishend
We empirically show that the $\mathsf{BPA}$ scheme can guarantee the context property that there exists a maximum of one pun residing in the text.

\begin{figure}
	\centering
	\includegraphics[width=0.9\linewidth]{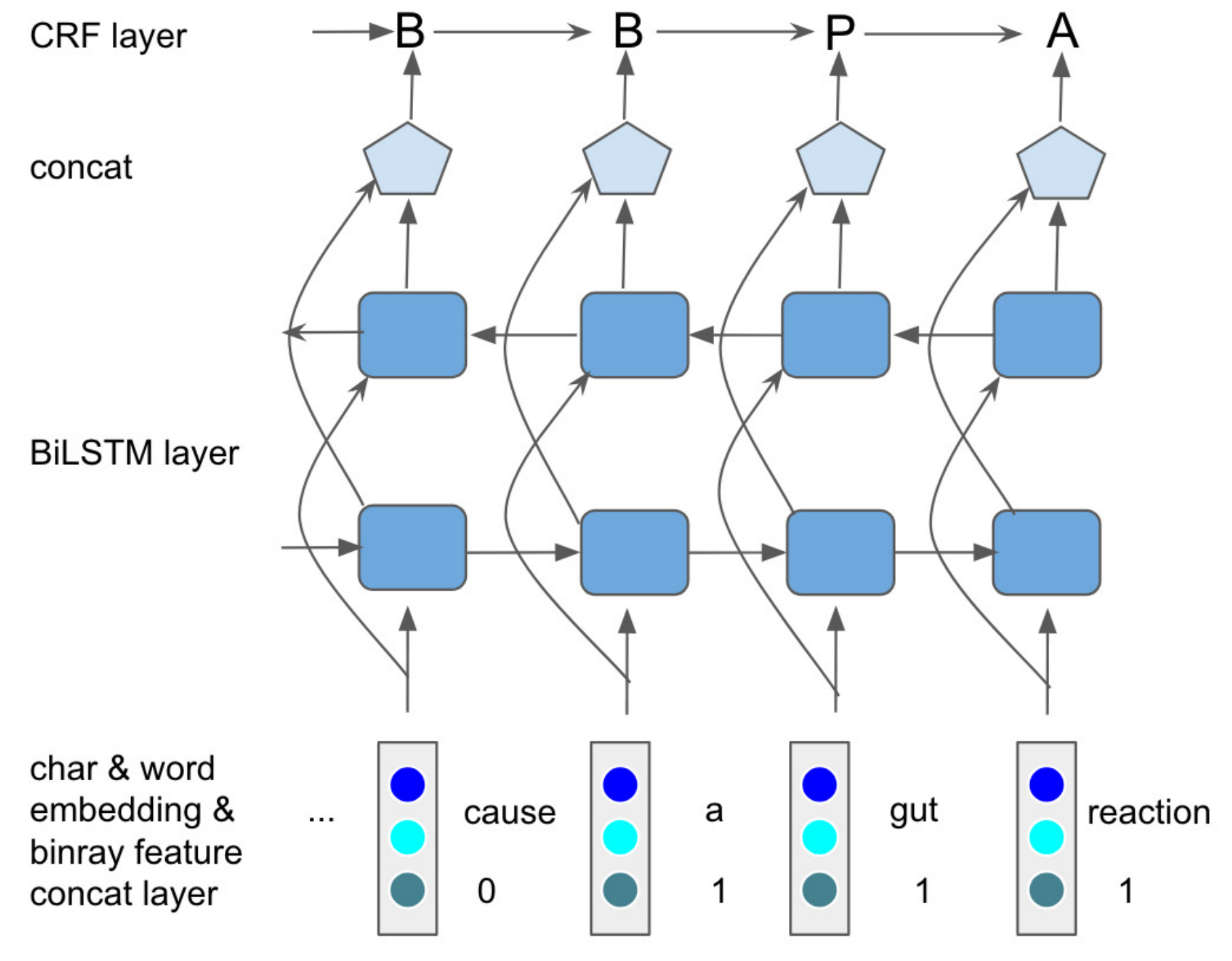}
	\vspace{-1mm}
	\caption{Model architecture}
	\label{fig:model}
	\vspace{-5mm}
\end{figure}

Given a context from the training set, we will be able to generate its corresponding gold tag sequence using a deterministic procedure.
Under the two schemes, if a sentence does not contain any puns, all words will be tagged with  $\mathsf{N}$ or $\mathsf{B}$, respectively.
Exemplified by the second sentence ``{\em Some diets cause a gut reaction}," the pun is given as ``{\em gut}."
Thus, under the $\mathsf{BPA}$ scheme, it should be tagged with $\mathsf{P}$, while the words before it are assigned with the tag $\mathsf{B}$ and words after it are with $\mathsf{A}$, as illustrated in Figure \ref{fig:model}.
Likewise, the $\mathsf{NP}$ scheme tags the word ``{\em gut}" with $\mathsf{P}$, while other words are tagged with $\mathsf{N}$.
Therefore, we can combine the pun detection and location tasks into one problem which can be solved by the sequence labeling approach.

\subsection{Model}
Neural models have shown their effectiveness on sequence labeling tasks \cite{chiu2015named,ma2016end,liu2018empower}.
In this work, we adopt the bidirectional Long Short Term Memory (BiLSTM) \cite{graves2013speech} networks on top of the Conditional Random Fields \cite{lafferty2001conditional} (CRF) architecture to make labeling decisions, which is one of the classical models for sequence labeling.
Our model architecture is illustrated in Figure \ref{fig:model} with a running example.
Given a context/sentence $\mathbf{x}=(x_1, x_2, \dots, x_n)$ where $n$ is the length of the context, we generate the corresponding tag sequence $\mathbf{y}=(y_1, y_2, \dots, y_n)$ based on our designed tagging schemes and the original annotations for pun detection and location provided by the corpora.
Our model is then trained on pairs of $(\mathbf{x},\mathbf{y})$.

\noindent
\textbf{Input. }
The contexts in the pun corpus hold the property that each pun contains exactly one content word, which can be either a noun, a verb, an adjective, or an adverb.
To capture this characteristic, we consider lexical features at the character level.
Similar to the work of \cite{liu2018empower}, the character embeddings are trained by the character-level LSTM networks on the unannotated input sequences.
Nonlinear transformations are then applied to the character embeddings by highway networks \cite{srivastava2015highway}, which map the character-level features into different semantic spaces.

We also observe that a pun tends to appear at the end of a sentence.
Specifically, based on the statistics, we found that sentences with a pun that locate at the second half of the text account for around 88\% and 92\% in homographic and heterographic datasets, respectively.
We thus introduce a binary feature that indicates if a word is located at the first or the second half of an input sentence to capture such positional information.
A binary indicator can be mapped to a vector representation using a randomly initialized embedding table \cite{P17-1044,bailin-lu:2018:AAAI2018}.
In this work, we directly adopt the value of the binary indicator as part of the input.

The concatenation of the transformed character embeddings, the pre-trained word embeddings \cite{pennington2014glove}, and the position indicators are taken as input of our model\footnote{
	The word sense has also been shown helpful for the location of a homographic pun \cite{cai2018sense}.
	However, such information may not always be helpful for the location of heterographic puns.
	We thus exclude such knowledge.}.

\begin{table*}[th]
	\centering
	\scalebox{0.70}{
		\begin{tabular}{l||ccc|ccc||ccc|ccc}
			\multirow{3}{*}{System}    & \multicolumn{6}{c||}{Homographic} & \multicolumn{6}{c}{Heterographic} \\ \cline{2-13}
			& \multicolumn{3}{c|}{Detection} & \multicolumn{3}{c||}{Location} & \multicolumn{3}{c|}{Detection} & \multicolumn{3}{c}{Location} \\
			\cline{2-13}
			&  \multicolumn{1}{c}{\em P.} & \multicolumn{1}{c}{\em R.}  & \multicolumn{1}{c}{\em F$_1$}  & \multicolumn{1}{|c}{\em P.} & \multicolumn{1}{c}{\em R.}  & \multicolumn{1}{c||}{\em F$_1$}  &  \multicolumn{1}{c}{\em P.} & \multicolumn{1}{c}{\em R.}  & \multicolumn{1}{c}{\em F$_1$}  & \multicolumn{1}{|c}{\em P.} & \multicolumn{1}{c}{\em R.}  & \multicolumn{1}{c}{\em F$_1$}  \\ \hline
			\hline
			\citet{pedersen2017duluth} & 78.32 & 87.24 & 82.54 & 44.00 & 44.00 & 44.00 & 73.99 & 86.62 & 68.71 & - & - & - \\  
			\citet{pramanick2017ju} & 72.51 & 90.79 & 68.84 & 33.48 & 33.48 & 33.48 & 73.67 & 94.02 & 71.74 & 37.92 & 37.92 & 37.92 \\  
			\citet{mikhalkova2017punfields}  & 79.93 & 73.37 & 67.82 & 32.79 & 32.79 & 32.79 & 75.80 & 59.40 & 57.47 & 35.01 & 35.01 & 35.01 \\  
			\citet{vadehra2017uwav} & 68.38 & 47.23 & 46.71 & 34.10 & 34.10 & 34.10 & 65.23 & 41.78 & 42.53 & 42.80 & 42.80 & 42.80 \\  
			\citet{indurthi2017fermi} & 90.24 & 89.70 & 85.33 & 52.15 & 52.15 & 52.15 & - & - & - & - & - & - \\ 
			\citet{vechtomova2017uwaterloo} & - & - & - & 65.26 & 65.21 & 65.23 & - & - & - & 79.73 & \textbf{79.54} & \textbf{79.64} \\  
			\citet{cai2018sense}& - & - & - & 81.50 & 74.70 & 78.00 & - & - & - & - & - & - \\ \hline
			\hline
			CRF & 87.21 & 64.09 & 73.89 & \textbf{86.31} & 55.32 & 67.43 & \textbf{89.56} & 70.94 & 79.17 & \textbf{88.46} & 62.76 & 73.42 \\ \hdashline
			Ours -- $\mathsf{NP}$ & 89.19 & 86.25 & 87.69 & 82.11 & 70.82 & 76.04 & 85.33 & 90.64 & 87.91 & 79.17 & 71.76 & 75.28 \\
			Ours -- $\mathsf{BPA}$ & 89.24 & 92.28 & 91.04 & 83.55 & \textbf{77.10} & \textbf{80.19} & 84.62 & \textbf{95.20} & 89.60 & 81.41 & 77.50 & 79.40 \\
			
			Ours -- $\mathsf{BPA}$-$\mathsf{p}$ & \textbf{91.25} & \textbf{93.28} & \textbf{92.19} & 82.06 & 76.54 & 79.20 & 86.67 & 93.08 & \textbf{89.76} & 80.81 & 75.22 &77.91 \\
			\hdashline 
			Pipeline & - & - & - & 67.70 & 67.70 & 67.70 & - & - & - & 68.84 & 68.84 & 68.84 \\ 
			
		\end{tabular}
	}
	\vspace{-2mm}
	\caption{Comparison results on two benchmark datasets. ({\em P.}: Precision, {\em R.}: Recall, {\em F$_1$}: {\em F}$_1$ score.)}
	\label{tab:result}
	\vspace{-5mm}
\end{table*}

\noindent
\textbf{Tagging.}
The input is then fed into a BiLSTM network, which will be able to capture contextual information.
For a training instance $(\mathbf{x}, \mathbf{y})$, we suppose the output by the word-level BiLSTM is $\mathbf{Z} = (\mathbf{z}_1, \mathbf{z}_2, \dots, \mathbf{z}_n)$.
The CRF layer is adopted to capture label dependencies and make final tagging decisions at each position, which has been included in many state-of-the-art sequence labeling models \cite{ma2016end,liu2018empower}.
The conditional probability is defined as:
\begin{center}
	\vspace{2.5mm}
	\begin{tabular}{c}
		$P(\mathbf{y}|\mathbf{x})
		=\frac{\prod_{i=1}^{n} \exp{(W_{y_{i-1},y_i}\mathbf{z}_i+b_{y_{i-1},y_i})}}{\sum_{\mathbf{y}'\in \mathbf{Y}}\prod_{i=1}^{n}\exp {(W_{y'_{i-1},y'_i}\mathbf{z}_i+b_{y'_{i-1},y'_i})}}$
	\end{tabular}
	\vspace{2.5mm}
\end{center}
where $\mathbf{Y}$ is a set of all possible label sequences consisting of tags from $\mathsf{\{N, P\}}$ (or $\mathsf{\{B, P, A\}}$), $W_{y_{i-1},y_i}$ and $b_{y_{i-1},y_i}$ are weight and bias parameters corresponding to the label pair $(y_{i-1},y_i)$.
During training, we minimize the negative log-likelihood summed over all training instances:
\begin{center}
	\begin{tabular}{c}
		$\mathcal{L}=-\sum_{i}\log P(\mathbf{y}_i|\mathbf{x}_i)$
	\end{tabular}
\end{center}
where $(\mathbf{x}_i,\mathbf{y}_i)$ refers to the $i$-th instance in the training set.
During testing, we aim to find the optimal label sequence for a new input $\mathbf{x}$:
\begin{center}
	\begin{tabular}{c}
		$\mathbf{y^\ast}
		=\mathop{\arg\max}_{\mathbf{y}\in \mathbf{Y}} P(\mathbf{y}|\mathbf{x})$
	\end{tabular}
\end{center}

This search process can be done efficiently using the Viterbi algorithm.

\section{Experiments}

\subsection{Datasets and Settings}
We evaluate our model on two benchmark datasets \cite{miller2017semeval}.
The homographic dataset contains 2,250 contexts, 1,607 of  which contain a pun.
The heterographic dataset consists of 1,780 contexts with 1,271 containing a pun.
We notice there is no standard splitting information provided for both datasets.
Thus we apply 10-fold cross validation.
To make direct comparisons with prior studies, following \cite{cai2018sense}, we accumulated the predictions for all ten folds and calculate the scores in the end.

For each fold, we randomly select 10\% of the instances from the training set for development.
Word embeddings are initialized with the 100-dimensional Glove \cite{pennington2014glove}.
The dimension of character embeddings is 30 and they are randomly initialized, which can be fine tuned during training.
The pre-trained word embeddings are not updated during training.
The dimensions of hidden vectors for both char-level and word-level LSTM units are set to 300.
We adopt stochastic gradient descent (SGD) \cite{bottou1991stochastic} with a learning rate of 0.015.

For the pun detection task, if the predicted tag sequence contains at least one $\mathsf{P}$ tag, we regard the output (i.e., the prediction of our pun detection model) for this task as true, otherwise false.
For the pun location task, a predicted pun is regarded as correct if and only if it is labeled as the gold pun in the dataset.
As to pun location, to make fair comparisons with prior studies, we only consider the instances that are labeled as the ones containing a pun.
We report precision, recall and $F_1$ score in Table \ref{tab:result}.
A list of prior works that did not employ joint learning are also shown in the first block of Table \ref{tab:result}.

\subsection{Results}
\label{sec:exp}
We also implemented a baseline model based on conditional random fields (CRF), where features like POS tags produced by the Stanford POS tagger \cite{toutanova2003feature}, n-grams, label transitions, word suffixes and relative position to the end of the text are considered.
We can see that our model with the $\mathsf{BPA}$ tagging scheme
yields new state-of-the-art $F_1$ scores on pun detection and competitive results on pun location, compared to baselines that do not adopt joint learning in the first block.
For location on heterographic puns, our model's performance is slightly lower than the system of \cite{vechtomova2017uwaterloo}, which is a rule-based locator.
Compared to CRF, we can see that our model, either with the $\mathsf{NP}$ or the $\mathsf{BPA}$ scheme, yields significantly higher recall on both detection and location tasks, while the precisions are relatively close.
This demonstrates the effectiveness of BiLSTM, which learns the contextual features of given texts -- such information appears to be helpful in recalling more puns.

Compared to the $\mathsf{NP}$ scheme, the $\mathsf{BPA}$ tagging scheme is able to yield better performance on these two tasks.
After studying outputs from these two approaches, we found that one leading source of error for the $\mathsf{NP}$ approach is that there exist more than one words in a single instance that are assigned with the $\mathsf{P}$ tag.
However, according to the description of pun in \cite{miller2017semeval}, each context contains a maximum of one pun.
Thus, such a useful structural constraint is not well captured by the simple approach based on the $\mathsf{NP}$ tagging scheme.
On the other hand, by applying the $\mathsf{BPA}$ tagging scheme, such a constraint is properly captured in the model.
As a result, the results for such a approach are significantly better than the approach based on the $\mathsf{NP}$ tagging scheme, as we can observe from the table.
Under the same experimental setup, we also attempted to exclude word position features.
Results are given by $\mathsf{BPA}$-$\mathsf{p}$.
It is expected that the performance of pun location drops, since 
such position features are able to capture the interesting property that a pun tends to appear in the second half of a sentence.
While such knowledge is helpful for the location task,
interestingly, a model without position knowledge yields improved performance on the pun detection task.
One possible reason is that detecting whether a sentence contains a pun is not concerned with such word position information.


Additionally, we conduct experiments over sentences containing a pun only, namely 1,607 and 1,271 instances from homographic and heterographic pun corpora separately.
It can be regarded as a ``pipeline'' method where the classifier for pun detection is regarded as perfect.\footnote{Under a pipeline setting, the first step is to detect if a sentence contains a pun. Then another algorithm is called to locate the exact pun word residing in the sentence if such a sentence is detected as the one containing a pun. In our setting, we assume the detection phase is perfect. In other words, all sentences containing a pun are exactly retrieved.}
Following the prior work of \cite{cai2018sense}, we apply 10-fold cross validation.
Since we are given that all input sentences contain a pun, we only report accumulated results on pun location, denoted as Pipeline in Table \ref{tab:result}.
Compared with our approaches, the performance of such an approach drops significantly.
On the other hand, such a fact demonstrates that the two task, detection and location of puns, can reinforce each other.
These figures demonstrate the effectiveness of our sequence labeling method to detect and locate English puns in a joint manner.

\subsection{Error Analysis}
We studied the outputs from our system and make some error analysis.
We found the errors can be broadly categorized into several types, and we elaborate them here.
1) Low word coverage:
since the corpora are relatively small, there exist many unseen words in the test set.
Learning the representations of such unseen words is challenging, which affects the model's performance.
Such errors contribute around 40\% of the total errors made by our system.
2) Detection errors: we found many errors are due to the model's inability to make correct pun detection. Such inability harms both pun detection and pun location. 
Although our approach based on the $\mathsf{BPA}$ tagging scheme yields relatively higher scores on the detection task, we still found that 40\% of the incorrectly predicted instances fall into this group.
3) Short sentences: we found it was challenging for our model to make correct predictions when the given text is short.
Consider the example ``{\em Superglue! Tom rejoined,}" here the word {\em rejoined} is the corresponding pun.
However, it would be challenging to figure out the pun with such limited contextual information.


\section{Related Work}
\label{sec:related}
Most existing systems address pun detection and location separately.
\citet{pedersen2017duluth} applied word sense knowledge to conduct pun detection.
\citet{indurthi2017fermi} trained a bidirectional RNN classifier for detecting homographic puns.
Next, a knowledge-based approach is adopted to find the exact pun.
Such a system is not applicable to heterographic puns.
\citet{doogan2017idiom} applied 
Google n-gram and word2vec to make decisions.
The phonetic distance via the CMU Pronouncing Dictionary is computed to detect heterographic puns.
\citet{pramanick2017ju} used the hidden Markov model and a cyclic dependency network with rich features to detect and locate puns.
\citet{mikhalkova2017punfields} used a supervised approach to pun detection and a weakly supervised approach to pun location based on the position within the context and part of speech features.
\citet{vechtomova2017uwaterloo} proposed a rule-based system for pun location that scores candidate words according to eleven simple heuristics.
Two systems are developed to conduct detection and location separately in the system known as {UWAV} \cite{vadehra2017uwav}.
The pun detector combines predictions from three classifiers.
The pun locator considers word2vec similarity between every pair of words in the context and position to pinpoint the pun.
The state-of-the-art system for homographic pun location is a neural method \cite{cai2018sense}, where the word senses are incorporated into a bidirectional LSTM model.
This method only supports the pun location task on homographic puns.
Another line of research efforts related to this work is sequence labeling, such as POS tagging, chunking, word segmentation and NER.
The neural methods have shown their effectiveness in this task, such as BiLSTM-CNN \cite{chiu2015named}, GRNN \cite{xu2016dependency}, LSTM-CRF \cite{lample2016neural}, LSTM-CNN-CRF \cite{ma2016end}, LM-LSTM-CRF \cite{liu2018empower}.

In this work, we combine pun detection and location tasks as a single sequence labeling problem.
Inspired by the work of \cite{liu2018empower}, we also adopt a LSTM-CRF with character embeddings to make labeling decisions.

\section{Conclusion}
In this paper, we propose to perform pun detection and location tasks in a joint manner from a sequence labeling perspective.
We observe that each text in our corpora contains a maximum of one pun.
Hence, we design a novel  tagging scheme  to incorporate such a constraint.
Such a scheme guarantees that there is a maximum of one word that will be tagged as a pun during the testing phase.
We also found the interesting structural property such as the fact that most puns tend to appear at the second half of the sentences can be helpful for such a task, but was not explored in previous works.
Furthermore, unlike many previous approaches, our approach, though simple, is generally applicable to both 
heterographic and homographic puns.
Empirical results on the benchmark datasets prove the effectiveness of the proposed approach that the two tasks of pun detection and location can be addressed by a single model from a sequence labeling perspective.

Future research includes the investigations on how to make use of richer semantic and linguistic information for detection and location of puns.
Research on puns for other languages such as Chinese is still under-explored, which could also be an interesting direction for our future studies.


\section*{Acknowledgments}
We would like to thank the three anonymous reviewers for their thoughtful and constructive comments.
This work is supported by Singapore Ministry of Education Academic Research Fund (AcRF) Tier 2 Project MOE2017-T2-1-156, and is partially supported by SUTD project PIE-SGP-AI-2018-01.

\bibliography{pun}
\bibliographystyle{acl_natbib}

\end{document}